\documentclass[10pt,twocolumn,letterpaper]{article}

\usepackage{cvpr}              
\usepackage[accsupp]{axessibility}  
\usepackage{graphicx}
\usepackage{amsmath}
\usepackage{amssymb}
\usepackage{booktabs}

\usepackage{times}
\usepackage{epsfig}
\usepackage{graphicx}
\usepackage{amsmath}
\usepackage{amssymb}
\usepackage{amsfonts}
\usepackage{float}

\usepackage{cite}

\usepackage{color}
\usepackage{colortbl}
\usepackage{multirow}
\usepackage{makecell}
\usepackage{hhline}
\usepackage{bbding}
\usepackage{booktabs}

\usepackage{arydshln}
\usepackage{stfloats}

\usepackage{algorithm}
\usepackage{algorithmic}

\usepackage[pagebackref,breaklinks,colorlinks]{hyperref}

\usepackage[capitalize]{cleveref}
\crefname{section}{Sec.}{Secs.}
\Crefname{section}{Section}{Sections}
\Crefname{table}{Table}{Tables}
\crefname{table}{Tab.}{Tabs.}

\def\cvprPaperID{*****} 
\def\confName{CVPR}
\def\confYear{2022}

\definecolor{lightgray}{gray}{1.0}
\definecolor{tinygray}{gray}{.96}
\definecolor{lightgray}{gray}{.6}

\newcommand{\notsure}[1]{\textcolor[rgb]{0,0,0}{#1}}

\def\cvprPaperID{9716} 
\def\confName{CVPR}
\def\confYear{2022}

\begin{document}
	
	\title{Semantic-shape Adaptive Feature Modulation for Semantic Image Synthesis}  
	\author{Zhengyao Lv$^1$, Xiaoming Li$^2$, Zhenxing Niu$^3$, Bing Cao$^4$, Wangmeng Zuo$^{2,5(}$\Envelope$^)$\\
		$^1$Tomorrow Advancing Life 
		$^2$Harbin Institute of Technology \\
		$^3$Machine Intelligence Lab, Alibaba Group  
		$^4$Tianjin University   
		$^5$Peng Cheng Laboratory \\
		{\tt\small \{cszy98, hit.xmshr\}@gmail.com wmzuo@hit.edu.cn}
	}
	
	\maketitle
	\thispagestyle{empty}
	
	\begin{abstract}
		Recent years have witnessed substantial progress in semantic image synthesis, it is still challenging in synthesizing photo-realistic images with rich details. Most previous methods focus on exploiting the given semantic map, which just captures an object-level layout for an image. Obviously, a fine-grained part-level semantic layout will benefit object details generation, and it can be roughly inferred from an object's shape. In order to exploit the part-level layouts, we propose a Shape-aware Position Descriptor (SPD) to describe each pixel's positional feature, where object shape is explicitly encoded into the SPD feature. Furthermore, a Semantic-shape Adaptive Feature Modulation (SAFM) block is proposed to combine the given semantic map and our positional features to produce adaptively modulated features. Extensive experiments demonstrate that the proposed SPD and SAFM significantly improve the generation of objects with rich details. Moreover, our method performs favorably against the SOTA methods in terms of quantitative and qualitative evaluation. The source code and model are available at \href{https://github.com/cszy98/SAFM}{SAFM}.
	\end{abstract}
	
	\section{Introduction}
	Semantic image synthesis is a kind of conditional image generation task, which aims to generate semantically aligned and photo-realistic images with the given semantic maps. Compared to unconditional image generation, it has significant flexibility in image generation since we can flexibly control the generated image content by drawing or editing the input semantic maps. Semantic image synthesis has been widely used in many practical scenarios, \eg, content creation and image editing~\cite{chen2017photographic,park2019semantic,zhu2020sean,tang2020local}. 
	
	Recently, Generative Adversarial Networks (GANs)~\cite{goodfellow2014generative} are broadly adopted to solve this problem and achieve impressive results. Most works attempt to model the mapping between different semantic classes and visual appearances. Park~\etal~\cite{park2019semantic} propose to use spatially-adaptive transformations (SPADE) learned from the input semantic layouts to modulate the activations in the generator. CC-FPSE~\cite{liu2019learning} subsequently extends SPADE by predicting spatially-varying conditional convolution kernels from the semantic layouts. Most recently, SC-GAN~\cite{wang2021image} exploits the learned semantic vectors to get spatially-variant and appearance-correlated convolution kernels and normalization parameters for the semantic stylization.

	\begin{figure}[t]
		\vspace{-4pt}
		\centering
		\includegraphics[width=1.0\linewidth]{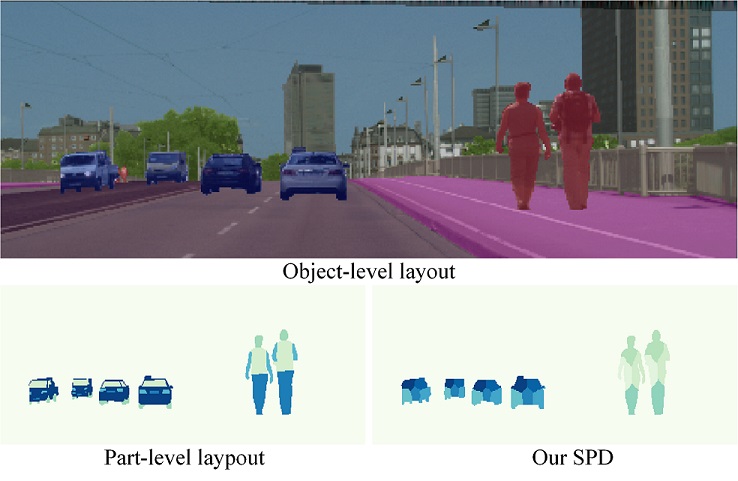}
		\vspace{-18pt}
		\caption{The given semantic map only provides an object-level layout, which is too coarse for generating images with rich details. The part-level semantic layout is implied in the shape/contour of an object instance. By encoding object shape into the proposed SPD feature, we can effectively exploit such part-level layouts for better image details generation.}
		\label{fig:cars}
		\vspace{-20pt}
	\end{figure}
	
	A semantic map has not only semantic labels but also a spatial layout. Such a spatial layout can be used to regularize the semantic image synthesis. Generally, one object instance is composed of some object parts, and pixels from the same object part should have a similar appearance while pixels from distinct object parts should not. For instance, an object `car' is composed of `window', `wheel', \etc Thus, the pixels from the `window' should look different from those from the `wheel'. In contrast, two pixels both from the `window' should look similar to each other. By exploiting such a spatial layout, we can suppress artifacts and generate coherent image details.    
	
	Semantic layouts have been effectively exploited to improve image synthesis in previous methods. However, the given semantic map just captures an \emph{object-level} layout for an image, which describes whether two pixels belong to the same object instance or not. It is too coarse to capture the fine-grained structure of an object instance. If we could subtly exploit a \textbf{part-level semantic layout}, it will benefit the generation of image high-frequency details.

	Obviously, the \emph{shape/contour} of each object instance can be easily identified from the object-level semantic layout. On the other hand, given the shape/contour of an object (\emph{e.g.,} a car), its part-level layout (\emph{e.g.,} the position of `window' or `wheel') can be roughly inferred according to the prior knowledge of an object's structure, as shown in Fig.~\ref{fig:cars}. Therefore, there is a strong connection between an object's shape and its part-level layout, \emph{i.e.,} 
	\textbf{an object's shape implies its part-level layout}. Thus, the exploitation of an object's part-level layout can be implicitly achieved by modeling its shape.
	
	In this paper, we propose a Shape-aware Position Descriptor (SPD) to describe each pixel's positional feature. Our SPD describes the relative relations (distance and angle) between each pixel inside an object instance and pixels on its contour, as shown in Figure.~\ref{fig:sc}~(a). Thus, the information of object shape has been encoded into each pixel's SPD feature. In other words, the clue of an object's part-level layout has been implicitly encoded into the SPD feature.
	
	Next, we design the Semantic-shape Adaptive Feature Modulation (SAFM) block to combine the given semantic map and our SPD features together, and modulate the input features adaptively. Specifically, our SAFM block first conditionally produces semantic-specific convolution kernels, and then performs semantic-specific convolution on the SPD features. At last, the SAFM block accepts input feature maps, adaptively modulates them, and forwards them to the next block, as shown in Figure.~\ref{fig:sc}~(b). 
	
	Note that our SPD is inspired by the shape context descriptor~\cite{belongie2000shape} which describes the relations of pixels just on the shape contour, but our SPD describes the relations between pixels inside an object and pixels on the contour.

	Our main contributions can be summarized as follows:
	\begin{itemize}
		\vspace{-4pt}
		\setlength{\itemsep}{0pt}
		\setlength{\parsep}{0pt}
		\item We propose a Shape-aware Position Descriptor (SPD) to describe the pixel's positional feature, where the object's part-level layout can be exploited and leveraged. 
		\item We design a Semantic-shape Adaptive Feature Modulation (SAFM) block, which combines the semantic maps and SPD features to produce adaptively modulated feature maps.
		\item Experimental results show that our method performs favorably on Cityscapes, COCO-stuff, and ADE20K datasets against SOTA methods and can generate more photo-realistic results with rich details.
	\end{itemize}

	\section{Related Work}
	
	\subsection{Semantic Image Synthesis}
	Generative Adversarial Networks (GANs)~\cite{goodfellow2014generative} have achieved impressive results on unconditional image generation related tasks~\cite{brock2018large,karras2019style,karras2020analyzing}. Subsequently, by introducing external information, such as class labels~\cite{mirza2014conditional,odena2016semi,odena2017conditional}, natural language descriptions~\cite{li2019controllable,li2020manigan,zhang2017stackgan}, or semantic maps~\cite{wang2018high,park2019semantic}, many kinds of conditional GANs are proposed to improve the controllability of image generation. 
	
	Semantic image synthesis is a task that takes semantic segmentation maps as input, which provides pixel-level class labels. Pix2Pix~\cite{isola2017image} is first proposed to use an encoder-decoder generator and PatchGAN discriminator to conduct semantic image generation. Pix2PixHD~\cite{wang2018high} improved it by adopting a coarse-to-fine generator and multi-scale discriminators to generate vivid details at high-resolution space. Particularly, Pix2PixHD introduced the instance-level boundary map as extra input to separate different instances for sharper boundaries. Further, panoptic aware convolutions and upsampling layers~\cite{dundar2020panoptic} are utilized to differentiate occluded instances.
	
	Recently, most works focus on how to sufficiently leverage the given semantic layouts. SPADE~\cite{park2019semantic} proposed to modulate the activations with spatially-adaptive transformations learned from semantic layouts. 
	CC-FPSE~\cite{liu2019learning} learned to predict conditional convolution kernels based on the given semantic layouts. Additionally, a feature pyramid semantic-embedding discriminator is employed to enable the generator to synthesize semantically aligned images with high-quality details. Similarly, Ntavelis~\etal~\cite{ntavelis2020sesame} proposed a two-stream discriminator by using semantic features to guide the scores of the discriminator. LGGAN~\cite{tang2020local} proposed a local class-specific and global image-level generative adversarial network to separately learn the global appearance distribution and generation of different object classes. EdgeGAN~\cite{tang2020edge} generated edges from semantic layouts to introduce detailed structure information for image synthesis. Most recently, SCGAN~\cite{wang2021image} learned semantic vectors to parameterize spatially conditional convolution and normalization. Besides, OASIS~\cite{sushko2020you} re-designed the discriminator with a segmentation-based network for synthesizing semantically aligned images with higher fidelity.
	
	Except for these GAN-based methods, CRN~\cite{chen2017photographic} adopted a cascaded refinement network for semantic image synthesis. Qi~\etal~\cite{qi2018semi} proposed a semi-parametric approach, which retrieved compatible fragments and composited them to assist the semantic image synthesis.
	 
	Most of those methods just exploit object-level semantic layouts, which are too coarse to capture the part-level structure of object instances. Although the part-level layouts are unknown, they can be roughly inferred from objects' shapes. In our work, such part-level layouts are effectively exploited by encoding objects' shapes into our SPD descriptors.

	\begin{figure*}[!t]
		\vspace{4pt}
		\centering
		\includegraphics[width=1.\linewidth]{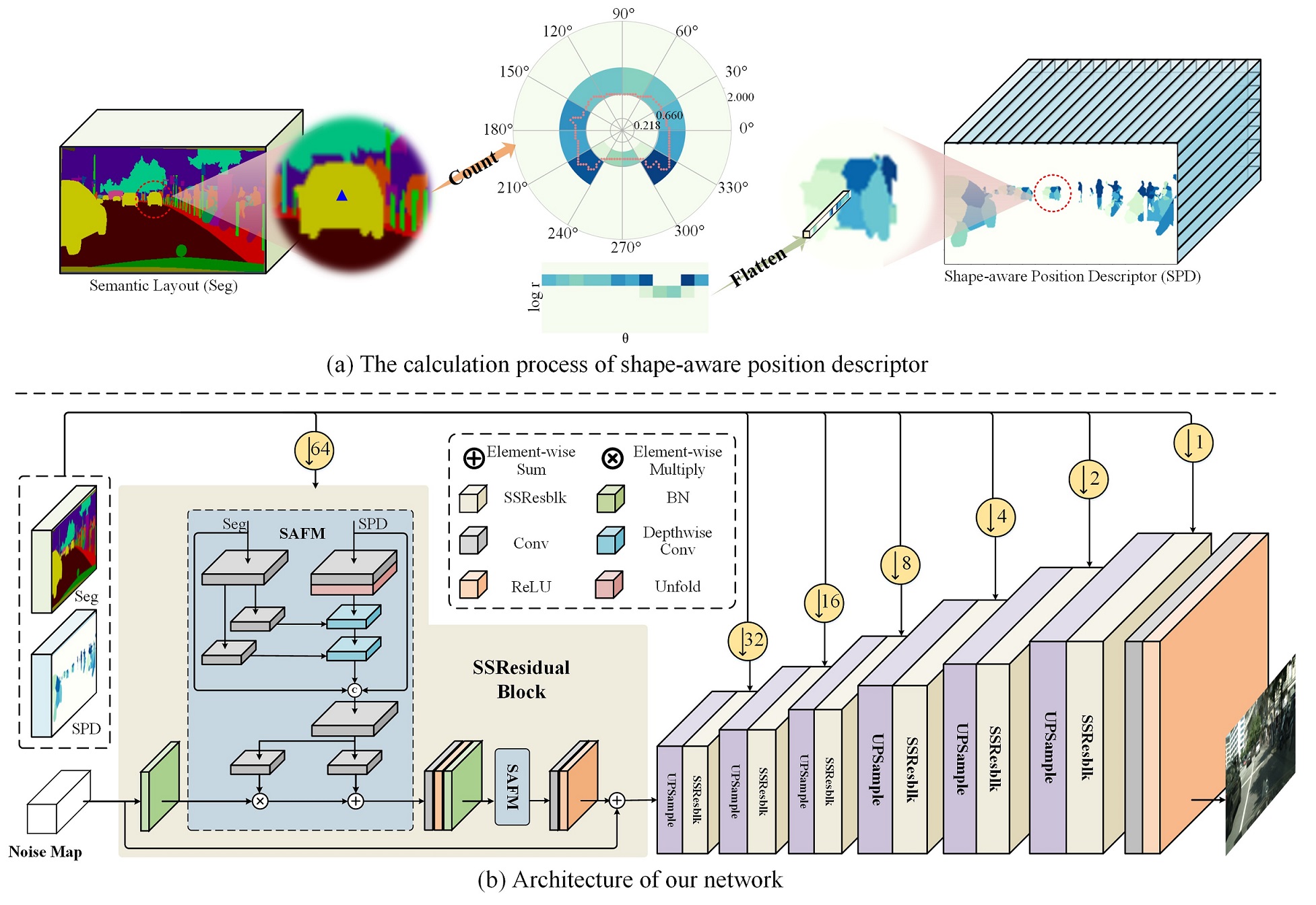}
		\caption{Overview of our proposed method. (a) shows the calculation process of the SPD features of a certain point in a car instance (denoted by a blue \textcolor{blue}{$\blacktriangle$}). After calculating all points inside the instances, we get a SPD map, as shown in (a) (right). (b) illustrates the architecture of our generator network, where SAFM is mainly constructed by conditional convolutions.}
		\label{fig:sc}
		\vspace{-10pt}
	\end{figure*}
	
	\subsection{Shape Context Descriptor}
	Shape context descriptor was first proposed by Belongie~\etal~\cite{belongie2000shape} for category-level shape matching and object recognition. 
	Through counting the histogram of the relative position distribution of other shape points, a rich local descriptor that implies the global shape points can be obtained for each point.
	After that, Thayananthan~\etal~\cite{thayananthan2003shape} propose an efficient dynamic programming scheme to constrain the figural continuity of shape context matching. Instead of Euclidean distance, Ling~\etal~\cite{ling2007shape} adopt inner-distance to measure the spatial relation between shape points, which can better capture the structure of complex shapes with articulations.
	
	The shape context descriptor can bring sufficient information that captures the relative locations within the whole instance beyond the point itself. 
	Due to the robustness and discrimination in reducing the ambiguity in class matching, these types of descriptors have been widely employed for different object recognition problems~\cite{belongie2002shape, mori2002estimating}, but are seldom exploited in semantic image synthesis tasks.
	In this work, we extend the shape context descriptor~\cite{belongie2000shape} to characterize the position of each point inside an object instance, where object shapes 
	are explicitly exploited and leveraged. 
	
	\section{Method}
	Given a semantic layout $S \in \mathbb{R}^{H \times W \times C}$ with $C$ class labels, our goal is to synthesize a photo-realistic image $I_s\in \mathbb{R}^{H \times W \times 3}$, which is semantically aligned with $S$. Following~\cite{wang2018high}, we adopt the instance-level segmentation map as supplementary input to obtain each instance region. 
	
	In the following, we first introduce the Shape-aware Position Descriptor (SPD), where object shapes are exploited and leveraged. Next, we design the Semantic-shape Adaptive Feature Modulation (SAFM) block to combine 
	the semantic maps and SPD features to adaptively modulate the input feature maps.

	\subsection{Shape-aware Position Descriptor}
	The shape of an object instance implies its part-level layout, as shown in Fig.~\ref{fig:cars}. In our approach, we propose the Shape-aware Position Descriptor (SPD) to describe each pixel’s positional feature, where the object shape is explicitly considered. In this way, the clue of an object’s part-level layout could be effectively exploited and leveraged.
	
	To balance the computation cost and the robustness of the descriptor, we only use the contour point set of an object instance to describe its shape information, instead of using all the points inside the segmentation region. 
	
	\noindent\textbf{Calculation process of the SPD.} 
	Figure.~\ref{fig:sc}~(a) illustrates the calculation process of our proposed SPD. Taking the rear-view car in the semantic map as an example, we can easily get its contour shape $T\in \{0,1\}^{H\times W}$ according to its segmentation map, where the points on the contour are denoted by label $1$ and the rest are set to $0$.
	Further, the contour shape of the car can be discretely represented as a point set $P=\{(x,y)|T(x,y)=1\}$. 
	
	For any point $o=(x_o, y_o)$ inside the instance, we calculate its positional descriptor through the following steps. 
	Firstly, we take the point $o$ as the pole to construct a polar coordinate space around it. 
	And then we divide this coordinate into $m\times n$ bins $B$ with $m$ polar radius intervals and $n$ polar angle intervals (in this work $m=12$ and $n=6$). Each bin $B_{i,j}$ should satisfy the following condition:
	\begin{equation}
	\setlength{\abovedisplayskip}{5pt}
	\setlength{\belowdisplayskip}{5pt}
	B_{i,j}=\{(r,\theta)|r_{i-1}<=r<r_{i},\theta_{j-1}<=\theta<\theta_{j}\}\,.
	\end{equation}
	In order to make the descriptor more sensitive to the nearby points relative to the farther ones, we use bins that are uniform in log-polar space. 
	
	After that, the distance and angle distribution of each point in the contour point set $P$ relative to the pole $o$ can be formulated as $P'=\{(r_i,\theta_i)|_{i=1}^{|P|}\}$. 
	Finally, we count the number of the points in the contour point set $P'$ that fall in each bin $B_{i,j}$, denoted as $H_{i,j}$:
	\begin{equation}
	\setlength{\abovedisplayskip}{5pt}
	\setlength{\belowdisplayskip}{5pt}
	H_{i,j}=|\{p|p\in P', p\in B_{i,j} \}|,
	\end{equation}
	where $|\cdot|$ denotes the quantization operation.
	By integrating the number of contour points in all $m\times n$ bins and flattening it, we can get a vector $v_o\in \mathbb{R}^{m\times n}$ about point $o$ that stores the contour point distribution. 
	The final SPD $\hat{v}_o$  for point $o$ can be obtained through the normalization:
	\begin{equation}
	\setlength{\abovedisplayskip}{5pt}
	\setlength{\belowdisplayskip}{5pt}
	\hat{v}_o = \frac{v_o}{|P'|}.
	\end{equation}
	
	After calculating the descriptor for all the points inside each instance, we can get a SPD map that explicitly represents the detailed position for each point.
	
	\begin{figure}[t]
		\centering
		\includegraphics[width=1.\linewidth]{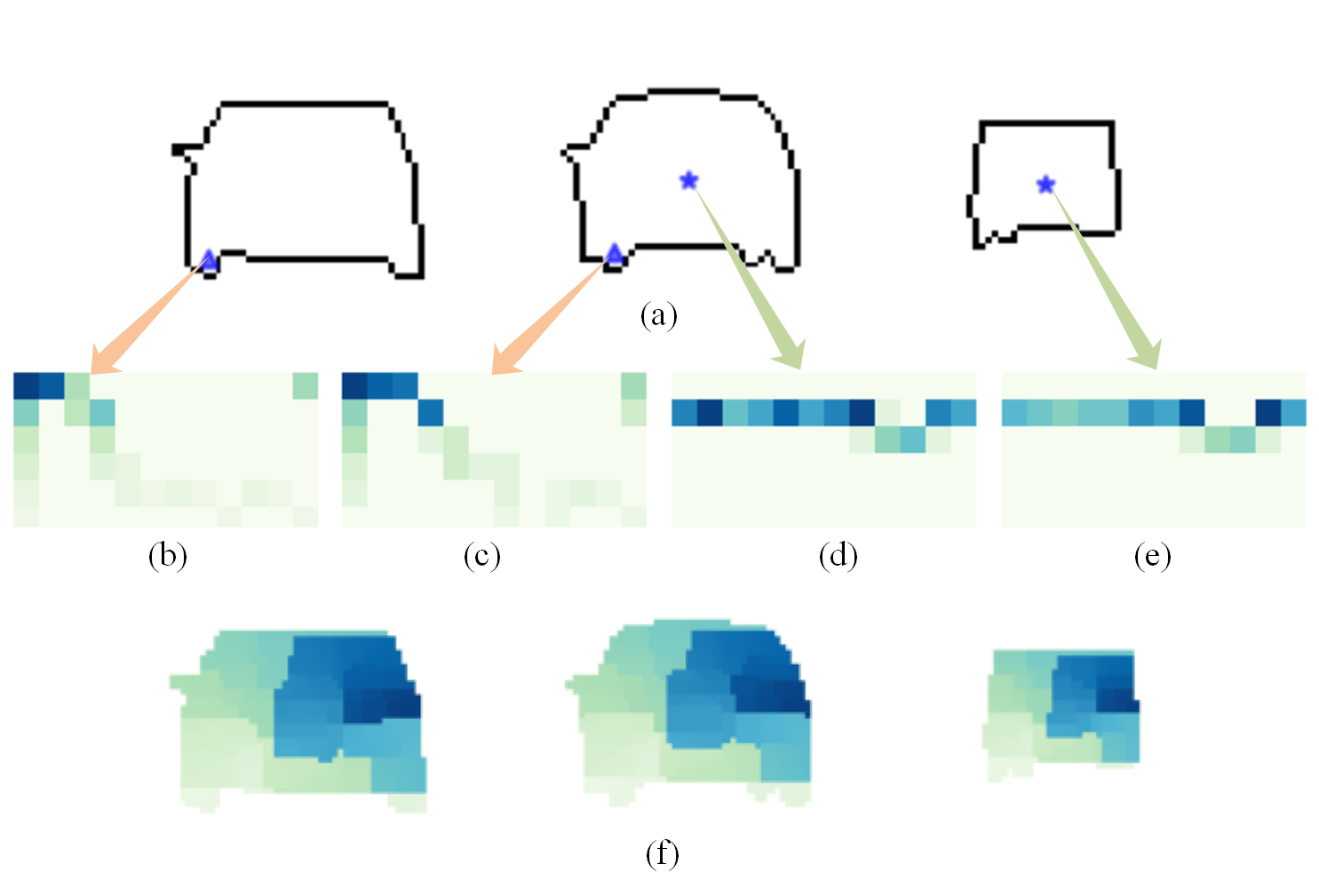}
		\vspace{-2pt}
		\caption{Visualization of the SPD features of the car instances. There are three different rear-view cars with similar shapes in (a). (b-e) show the descriptor of blue \textcolor{blue}{$\blacktriangle$} and \textcolor{blue}{$\star$} points in (a). (f) illustrates the descriptor compressed to 1D by t-SNE.}
		\label{fig:sean}
		\vspace{-8pt}
	\end{figure}
	
	\noindent\textbf{Discussion about the SPD.}
	For example, there are three rear-view cars in Fig. \ref{fig:sean} (a), which have similar shape contours but different spatial locations and scales. Intuitively, the corresponding points at the same object part should have consistent SPD features. In contrast, the points from different object parts should have different SPD features. 
	
	Taking the points on the left wheel as an example (denoted by blue  \textcolor{blue}{$\blacktriangle$}), the SPDs of the two points are shown in Fig. \ref{fig:sean} (b) and (c). 
	Another example is the points at  the center of car (denoted by blue \textcolor{blue}{$\star$}), their SPD features are shown in Fig. \ref{fig:sean} (d) and (e). 
	
	We can observe that: (i) For the corresponding positions of different instances, their SPD features look similar to each other (b \textit{vs} c and d \textit{vs} e). (ii) For different positions of the same instance, there are obvious differences between their SPD features (c \textit{vs} d). (iii) Even if the absolute location or scale of an instance changes, we can still get similar SPD features, which indicates our SPD is only dependent on object shape (d \textit{vs} e). 
	In other words, object shapes are robust and discriminatively encoded into our SPD features.
	
	Furthermore, we also jointly consider the SPD features of all pixels inside an object instance. Specifically, we use t-SNE to compress each pixel's SPD feature as a scalar, and hence we obtain a compact 2D map corresponding to 
	all pixels inside an object instance, as shown in (f). We can see that all the three cars share similar patterns in the compact 2D map. 
	More importantly, for each instance, \textbf{the compact 2D map could well describe the part-level layout of a car}. Thus, we claim that our SPD features could implicitly exploit an object's part-level layout.

	\subsection{Semantic-Shape Adaptive Feature Modulation}
\begin{figure}[t]
		\centering
		\includegraphics[width=1.\linewidth]{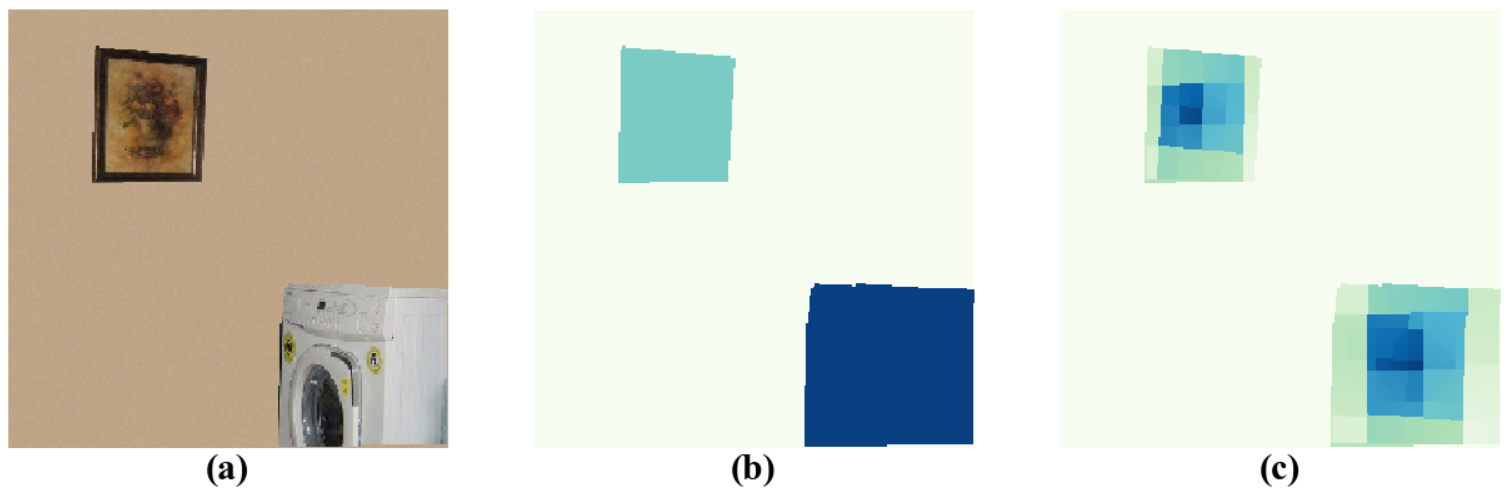}
		\vspace{-3pt}
		\caption{The (a) appearance, (b) shape and (c) the compact 2D map of the painting and washer instances.}
		\label{fig:sim}
	\end{figure}

	Another thing should be noticed is that different classes of object instances may have similar shapes.  
	For example, the painting and washer shown in Fig. \ref{fig:sim} (a) are both rectangular, and the patterns of their SPD features are quite similar (compact 2D map), but their appearance and structure are different, which will confuse the image synthesis.
	
	To circumvent this issue, we designed the Semantic-shape Adaptive Feature Modulation (SAFM) block (Fig. \ref{fig:sc} (b)), which combines the semantic information and our SPD features to compensate each other. For instance, the proposed SPD features can bring more detailed descriptions about the point positions to the semantic layouts, while the semantic layouts can bring complementary semantic information to the SPD features. And then the SAFM yields semantic-shape adaptive modulation parameters for different positions of different classes, so as to subtly guide the semantic image synthesis.
	
	In the SAFM block, semantic layouts and SPD features are first scaled to the same size. Then semantic layouts are fed into two convolution layers to predict two sets of semantic-adaptive $3\times3$ convolution kernels, which vary with the class label of spatial position. After that, through depthwise convolution layers, the semantics information of each spatial position is fused into the corresponding position in the SPD features. Finally, it yields semantic-shape adaptive modulation parameters with the fused features for feature modulation.
	
\begin{figure}[!t]
		\vspace{-2pt}
		\centering
		\includegraphics[width=1.\linewidth]{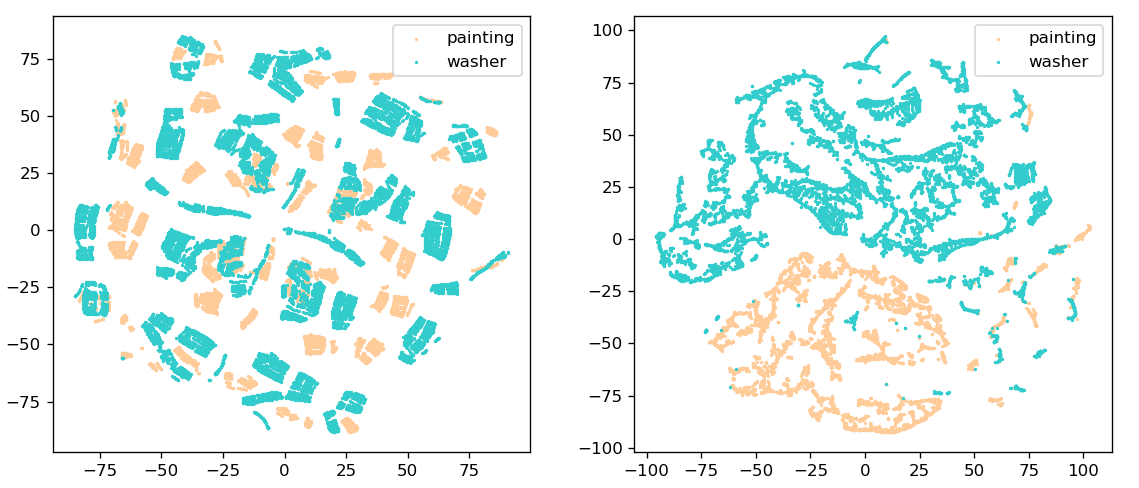}
		\vspace{-3pt}
		\caption{The distribution of the SPD of points in the painting and washer before and after using SAFM block.(mapping by t-SNE).}
		\label{fig:sep}
		\vspace{-10pt}
	\end{figure}

	With the SAFM block, semantic information and spatial position information can be integrated together. 
	Fig. \ref{fig:sep} shows the distribution of the SPD features without and with the SAFM block. Note that the light green dots represent the washer, while the orange dots stand for the painting. One can see that by incorporating the SAFM block, the washer and painting points tend to be better separated, which shows the effectiveness of the SAFM block in combining the SPD features and semantic features.
	
	\vspace{-8pt}
	\begin{table*}[t]
		\vspace{6pt}
		\begin{center}
			\small
			\renewcommand\arraystretch{1.}
			\setlength{\tabcolsep}{2.5mm}
			{
				\begin{tabular}{c c c c c c c c c c}
					\toprule[1.pt]
					& \multicolumn{3}{c}{\textbf{Cityscapes}} & \multicolumn{3}{c}{\textbf{ADE20K}} & \multicolumn{3}{c}{\textbf{COCO-Stuff}}\\
					\multirow{-2}{*}{\makecell[c]{\textbf{Methods}}} & \textbf{mIoU $\uparrow$} & \textbf{Acc $\uparrow$} & \textbf{FID $\downarrow$} & \textbf{mIoU $\uparrow$} & \textbf{Acc $\uparrow$} & \textbf{FID $\downarrow$} & \textbf{mIoU $\uparrow$} & \textbf{Acc $\uparrow$} & \textbf{FID $\downarrow$} \\
					\hline
					CRN~\cite{chen2017photographic}        & 52.4 & 77.1 & 104.7 & 22.4 & 68.8 & 73.3 & 23.7 & 40.4 & 70.4 \\
					SIMS~\cite{qi2018semi}        & 47.2 & 75.5 & 49.7 & N/A & N/A & N/A & N/A & N/A & N/A  \\
					pix2pixHD~\cite{wang2018high}       & 58.3   & 81.4  & 95.0  & 20.3  & 69.2  & 81.8  & 14.6 & 45.7 & 111.5\\
					SPADE~\cite{park2019semantic}    & 62.3 & 81.9 &  71.8 & 38.5 & 79.9 & 33.9 & 37.4 & 67.9 & 22.6  \\
					CC-FPSE~\cite{liu2019learning}  & 65.6 & 82.3 & 54.3 & 43.7 & 82.9 & 31.7 & 41.6 & 70.7 & 19.2  \\
					LGGAN~\cite{tang2020local} & 68.4 & 83.0 & 57.7 & 41.6 & 81.8 & 31.6 & N/A & N/A & N/A  \\
					OASIS~\cite{sushko2020you} & 69.3 & N/A & \textbf{47.7} & 48.3 & N/A & \textbf{28.3} & \textbf{44.1} & N/A & \textbf{17.0} \\
					SC-GAN~\cite{wang2021image} & 66.9 & 82.5 & 49.5 &45.2 &83.8 & 29.3 & 42.0 &72.0 &18.1  \\
					
					Ours & \textbf{70.4} & \textbf{83.1}  & 49.5 & \textbf{50.1} & \textbf{86.6} & 32.8 & 43.3 & \textbf{73.4} & 24.6 \\
					\bottomrule[1.pt]
				\end{tabular}
			}
		\end{center}
		\vspace{-12pt}
		\caption{The quantitative comparison with the competing methods on different datasets. $\uparrow$ ($\downarrow$) indicates higher (lower) is better.}
		\label{tab:quan}
		\vspace{-8pt}
	\end{table*}

	\subsection{Learning Objective}
	
	In our approach, we adopt adversarial loss $\mathcal{L}_{adv}$, feature matching loss $\mathcal{L}_{fm}$, and perceptual loss $\mathcal{L}_{perc}$ to achieve high fidelity and realness of generation. In addition, $\mathcal{L}_{seg}$ is suggested from the pre-trained segmentation model for constraining the semantic alignment.
	
	\noindent\textbf{Adversarial Loss.} Adversarial learning can effectively keep the generated images staying at the real image manifold and has been widely used in many image generation tasks~\cite{mirza2014conditional,brock2018large,karras2019style}. 
	In this work, we adopt the hinge-based adversarial loss~\cite{lim2017geometric,miyato2018spectral,zhang2019self} and the optimization of generator $G$ and discriminator $D$ can be formulated as: 
	\begin{equation}
	\setlength{\abovedisplayskip}{5pt}
	\setlength{\belowdisplayskip}{5pt}
	\begin{aligned}
	\mathcal{L}_{adv}^D = &-\mathbb{E}_{(I,S)}[min(0,-1+D(I,S))]
	\\-& \mathbb{E}_{(z,S)}[min(0,-1-D(G(z,S),S))],
	\end{aligned}
	\end{equation}
	
	\begin{equation}
	\setlength{\abovedisplayskip}{5pt}
	\setlength{\belowdisplayskip}{5pt}
	\begin{aligned}
	\mathcal{L}_{adv}^G = - \mathbb{E}_{(z,S)}D(G(z,S),S),
	\end{aligned}
	\end{equation}
	where $I$ is the real image, $S$ is the corresponding semantic layouts, and $z$ is the noise map fed into the generator.
	
	\noindent\textbf{Feature Matching Loss.} Following~\cite{wang2018high}, we adopt the feature matching loss $\mathcal{L}_{fm}$ to enhance the supervision for stabilizing the training process which constrains the features of synthesized images to be close to the real one in different feature spaces of discriminator $D$. This can be defined as:
	\begin{equation}
	\setlength{\abovedisplayskip}{5pt}
	\setlength{\belowdisplayskip}{5pt}
	\begin{aligned}
	\mathcal{L}_{fm} = \sum_{i=1}^n\frac{1}{N_i}||D_i(I,S)-D_i(G(z,S),S)||_1,
	\end{aligned}
	\end{equation}
	where $N_i$ is the number of elements in feature $D_i(I,S)$.
	
	\noindent\textbf{Perceptual Loss.}
	We adopt the pre-trained VGG19 model $\Phi$~\cite{simonyan2014very} to separately extract the features from the real image $I$ and the generated images $\hat{I}$. The perceptual loss $L_{perc}$ is computed in multi-scale feature space and is formulated as: 
	\begin{equation}
	\setlength{\abovedisplayskip}{5pt}
	\setlength{\belowdisplayskip}{5pt}
	\begin{aligned}
	\mathcal{L}_{perc} = \sum_{k=1}^K||\Phi_k(\hat{I})-\Phi_k(I)||_1,
	\end{aligned}
	\end{equation}
	where $\phi_k$ denotes the $k$-th feature map extracted from the VGG19 model $\Phi$. In our implementation, we set $K=5$.
	
	\noindent\textbf{Semantic Alignment Loss.} In order to explicitly constrain the semantic consistency between the generated image and the given semantic layout, we further introduce the semantic alignment loss $\mathcal{L}_{seg}$ to optimize the learning process:
	\begin{equation}
	\setlength{\abovedisplayskip}{5pt}
	\setlength{\belowdisplayskip}{5pt}
	\small
	\!\mathcal{L}_{seg}\!\!=\!\! -\!\!\sum_{i=1}^C\!\!w_i\!\!\sum_{j=1}^H\sum_{k=1}^W\!\!S_{i,j,k}[\log Seg(I)_{i,j,k}
	\!+\log Seg(\hat{I})_{i,j,k}],
	\end{equation}
	\begin{equation}
	\setlength{\abovedisplayskip}{5pt}
	\setlength{\belowdisplayskip}{5pt}
	\small
	w_i = \frac{H\times W}{\sum_{j=1}^H\sum_{k=1}^WS_{i,j,k}},
	\end{equation}
	where $Seg$ is a pre-trained segmentation  model ~\cite{badrinarayanan2017segnet}.
	
	The overall learning objective can be summarized as:
	\begin{equation}
	\setlength{\abovedisplayskip}{5pt}
	\setlength{\belowdisplayskip}{5pt}
	\mathcal{L} = \lambda_{adv}\mathcal{L}_{adv}^G+ \lambda_{fm}\mathcal{L}_{fm} +\lambda_{perc}\mathcal{L}_{perc} + \lambda_{seg}\mathcal{L}_{seg},
	\end{equation}
	where $\lambda_{adv}$, $\lambda_{fm}$, $\lambda_{perc}$, and $\lambda_{seg}$ are trade-off parameters.
	
	\section{Experiments}
	Extensive experiments are conducted to evaluate the effectiveness of our proposed SPD features and SAFM block. We report the quantitative and qualitative results in comparison with the competing methods, including CRN~\cite{chen2017photographic}, SIMS~\cite{qi2018semi}, Pix2PixHD~\cite{wang2018high}, SPADE~\cite{park2019semantic}, CC-FPSE~\cite{liu2019learning}, OASIS~\cite{sushko2020you}, LGGAN~\cite{tang2020local} and SC-GAN~\cite{wang2021image}. Besides, the ablation study is further conducted to explore the benefits of each component of our method that bring to the results.
	
	\subsection{Dataset and Experimental Details}
	
	\noindent\textbf{Dataset.} Our experiments are conducted on three challenging datasets, \ie, Cityscapes~\cite{cordts2016cityscapes}, ADE20K~\cite{zhou2017scene}, and COCO-Stuff~\cite{caesar2018coco}. The Cityscapes dataset contains images of urban street scenarios, which has 3,000 images for training and 500 for validation. The ADE20K dataset has 20,210 images for training and 2,000 for validation each of which has 150 semantic classes, covering indoor and outdoor scenarios. Similarly, The COCO-Stuff contains 182 classes that cover diverse scenarios and provides 118,000 images for training and 5,000 for validation. The real images and their corresponding semantic layouts in ADE20K and COCO-Stuff are resized and cropped to $256\times 256$ while those in Cityscapes are processed to $256\times 512$.
	\begin{figure*}[!t]
		\centering
		\includegraphics[width=1.\linewidth]{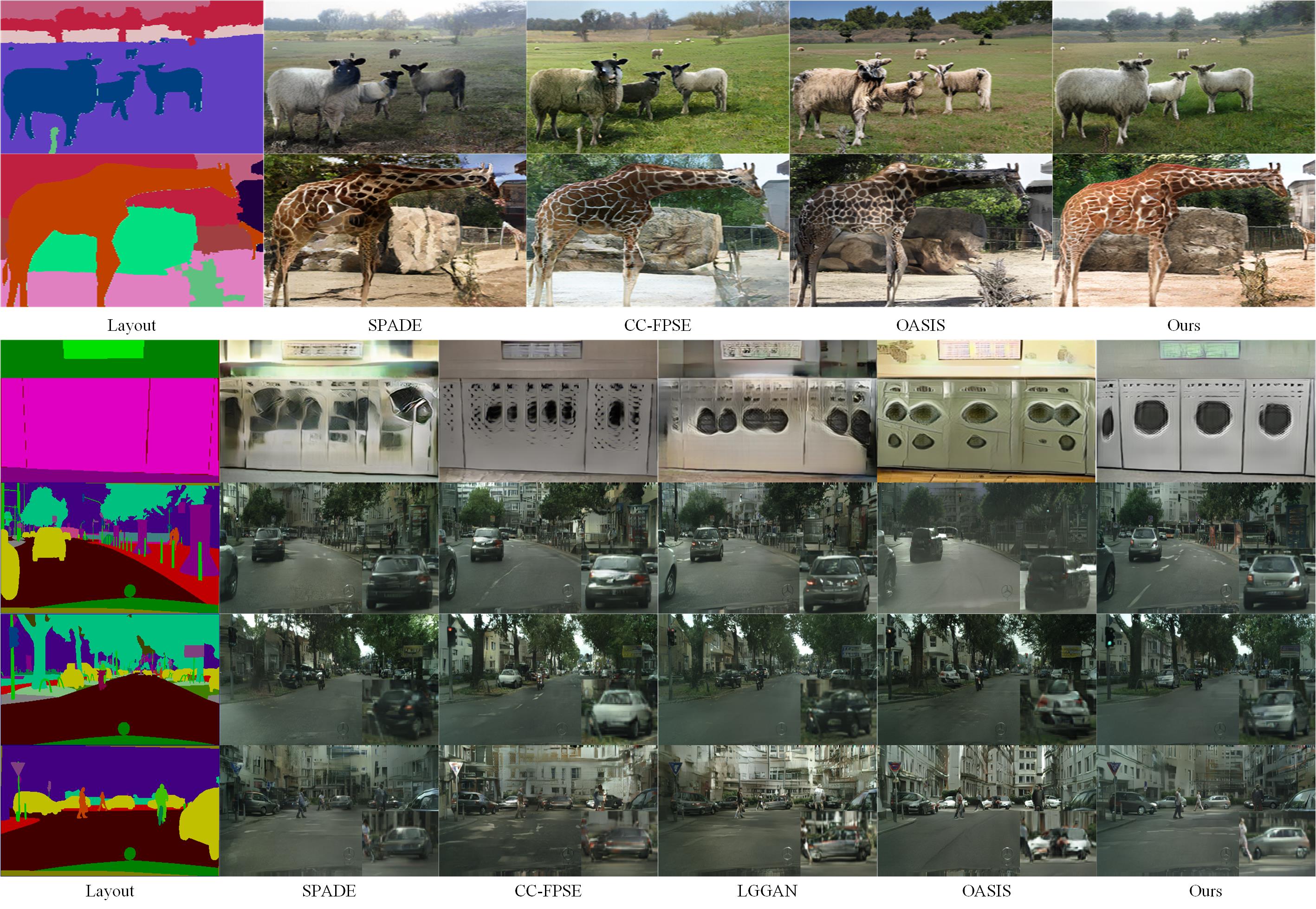}
		\vspace{-15pt}
		\caption{Visual comparisons on the COCO-Stuff ($1\textit{-}st \sim 2\textit{-}nd$ rows), ADE20K ($3\textit{-}rd$ row) and Cityscapes ($4\textit{-}th\sim 6\textit{-}th$ rows) datasets.}
		\label{fig:compare}
		\vspace{-5pt}
	\end{figure*}

	\noindent\textbf{Experimental Details.} 
	We adopt the generator of SPADE~\cite{park2019semantic} and SESAME~\cite{ntavelis2020sesame} discriminator as baseline model. Following SPADE, Spectral Norm~\cite{miyato2018spectral} is incorporated in all convolutional layers in our model.
	We adopt the ADAM~\cite{kingma2014adam} optimizer with $\beta_1=0, \beta_2=0.999$, and the learning rate is set to $1\times10^{-4}$ and $4\times10^{-4}$ for generator and discriminator, respectively. Our model is trained on ADE20K and Cityscapes for 200 epochs, and 100 epochs on COCO-Stuff. The trade-off parameters $\lambda_{adv}$, $\lambda_{fm}$, $\lambda_{perc}$, and $\lambda_{seg}$ are set to 1, 10, 10, 1, respectively.
	The experiments are carried out on a server with 4 2080Ti GPUs.
	
	\noindent\textbf{Evaluation Metrics.}
	Following previous semantic synthesis works~\cite{park2019semantic,liu2019learning}, we use three metrics to quantitatively evaluate the performance, \ie,  Fréchet Inception Distance (FID)~\cite{heusel2017gans}, mean Intersection-over-Union (mIoU), and pixel accuracy (Acc). Among these metrics, FID is introduced to assess the realism of the synthesized images by computing the Wasserstein-2 distance between the distributions of the synthesized and real images. Acc and mIoU are proposed to measure the differences of semantic labels between the synthesized images and the input semantic layouts. 
	Following~\cite{park2019semantic},  
	we use the pre-trained semantic segmentation models DRN-D-105~\cite{yu2017dilated}, UperUnet101~\cite{xiao2018unified} and DeepLabV2~\cite{chen2017deeplab} for the semantic evaluation of Cityscapes, ADE20K and COCO-Stuff, respectively. 
	In addition, we demonstrate the generated results for visual comparison with other competing methods. 
	Finally, a user study is reported to further evaluate the effectiveness of our method.
	\subsection{Quantitative and Qualitative Results}
	\noindent\textbf{Quantitative comparisons.} 
	Table \ref{tab:quan} lists the semantic segmentation and FID performance of our model and the competing methods on the Cityscapes, ADE20k, and COCO-Stuff datasets. In terms of semantic alignment, the mIoU of our method achieves 70.4 and 50.1 on Cityscapes and ADE20K, respectively (at least 1.1 and 1.8 higher than the second-best one). In addition, our method obtains the comparable FID scores, which ensures the distribution consistency between the generated results and the real images. The best semantic segmentation performance of our method indicates that the results of our method are not only more consistent with the target layout, but also photo-realistic in appearance, both of which can be attributed to the introduction of the SPD feature and the SAFM block.
	
	Note that OASIS achieves nearly the best performance on the COCO-Stuff dataset, but performs inferior to ours on the Cityscapes and ADE20K datasets, we analyze that the COCO-stuff dataset has more stuff classes without part-level semantics (91 stuff classes that cover about $66\%$ of the pixels), which makes the superiority of our SPD features for object instances not obvious in the quantitative results.
	
	\noindent\textbf{Qualitative comparisons.} 
	Fig.~\ref{fig:compare} gives the qualitative comparisons on the Cityscapes, ADE20K, and COCO-Stuff datasets, from which we can observe that (i) with the SPD, our method can generate more realistic details (\eg the washing machine in the 3\textit{-}$rd$ row), which is benefited from the discriminative and effective spatial position characterization. (ii) From the 4\textit{-}$th$ to 6\textit{-}$th$ rows, our method can well handle the instances of the same class with different shapes, while others fail to generate plausible results, indicating the effectiveness of our SPD features and SAFM block. (iii) With the constraints of semantic alignment, our method also performs well in unstructured textures, contributing to better visual quality, which can be seen from the $1\textit{-}st$ row. 
	
	\noindent\textbf{User Study.} Following \cite{park2019semantic}, we conduct a user study to further compare our method with SPADE, CC-FPSE, LGGAN, and OASIS on the Cityscapes and ADE20K datasets. For each set of experiments, participants with computer vision backgrounds\footnote{The participants have been informed that the collected data will be only used for academic purposes, and their identities will not be recorded.} are required to select the image that has better performance in semantic alignment and photo-realistic appearance. From Table \ref{tab:user} we can see that users are more likely to favor our results, especially on the Cityscapes.
	\begin{table}
		\begin{center}
			\small
			\renewcommand\arraystretch{1.1}
			\setlength{\tabcolsep}{5mm}
			{
				\begin{tabular}{c|c|c}
					\hline
					\textbf{Methods} & Cityscapes & ADE20K  \\
					\hline
					Ours $>$ SPADE & $74.76\%$ &  $63.32\%$  \\  
					Ours $>$ CC-FPSE& $63.20\%$ & $58.24\%$   \\ 
					Ours $>$ LGGAN& $68.48\%$ & $58.96\%$   \\ 
					Ours $>$ OASIS & $65.24\%$ & $56.76\%$   \\  
					\hline
				\end{tabular}
			}
		\end{center}
		\vspace{-16pt}
		\caption{User study. The numbers represent the percentage of our method favored by users relative to competing methods.}
		\label{tab:user}
		\vspace{-12pt}
	\end{table}
	
\begin{figure}[t]
		\vspace{-1pt}
		\centering
		\includegraphics[width=.97\linewidth]{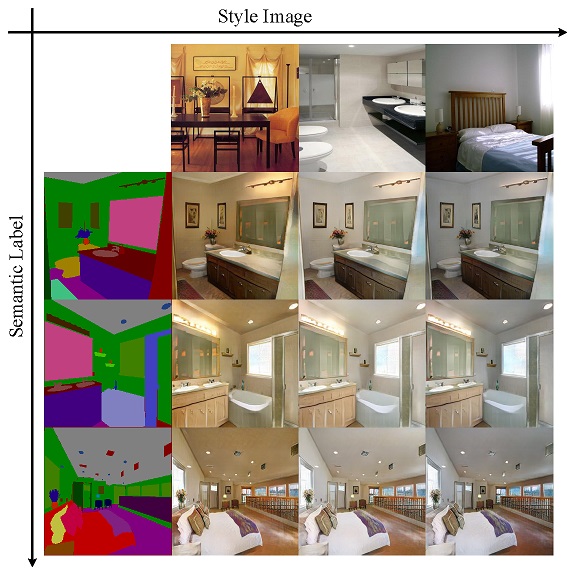}
		\vspace{-16pt}
		\caption{Visual results of multi-modal synthesis. }
		\label{fig:mmdal}
		\vspace{-6pt}
	\end{figure}

	\noindent\textbf{Multi-modal synthesis.} 
	Following SPADE~\cite{park2019semantic}, we train an additional encoder for multi-modal synthesis or style-guided image with the KL Divergence loss in the way of VAE~\cite{kingma2013auto}. By controlling the mean and variance vector to sample different random noises, our generator can also synthesize images with diverse and photo-realistic appearances for the given input segmentation mask, as shown in Fig.~\ref{fig:mmdal}.
	
	\noindent\textbf{Results with segmentation-based discriminator.} Noting the success of the segmentation-based discriminator in OASIS, we verified the effectiveness of SPD with the discriminator and training tricks of OASIS. Specifically, we use SPD to replace the 3D noise in the OASIS generator, resulting in improved results for the Cityscapes dataset (FID: 43.81 and mIoU: 71.8). More qualitative results are shown in the supplementary materials.
	
	\subsection{Ablation Study}
	
	\begin{figure*}[!t]
		\vspace{0pt}
		\centering
		\includegraphics[width=1.\linewidth]{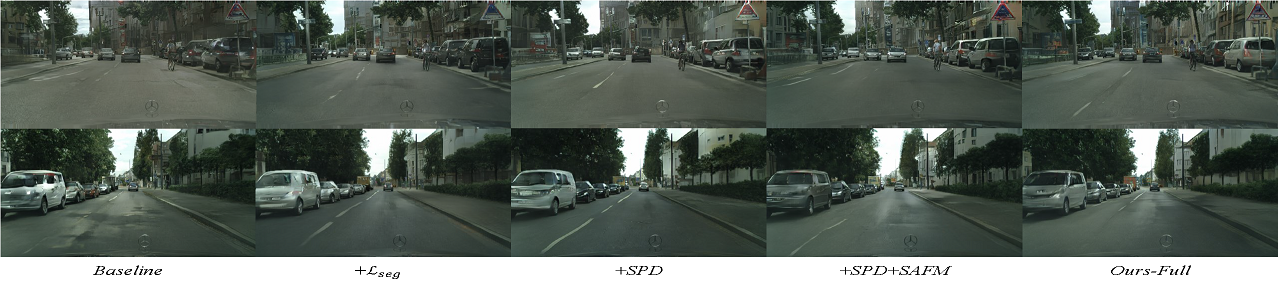}
		\vspace{-12pt}
		\caption{Visual comparisons of different variants.}
		\label{fig:ablation}
		\vspace{-8pt}
	\end{figure*}
	
	We conduct the ablation study on the Cityscapes dataset to evaluate the effectiveness of our SPD feature and SAFM block, which contains the following variants. 
	(1) \textit{Baseline}: by adopting SPADE~\cite{park2019semantic} generator and SESAME~\cite{ntavelis2020sesame} discriminator as the baseline model.
	(2) \textit{Baseline+}$\mathcal{L}_{seg}$: by adding the semantic alignment loss upon the \textit{Baseline} model. 
	(3) \textit{Baseline+SPD}: by concatenating the SPD features with semantic layouts and feeding them into the SPADE block to generate spatially adaptive modulation parameters. 
	(4) \textit{Baseline+SPD+SAFM}: by introducing the SAFM block to the generator to exploit the SPD features instead of directly concatenating them. 
	(5) \textit{Ours-Full}: by incorporating the Baseline, $\mathcal{L}_{seg}$, SPD features and SAFM block together. The quantitative results and visual comparisons are shown in Table~\ref{tab:ablation} and Fig.~\ref{fig:ablation}, respectively.
	
	We can see that (i) although $\mathcal{L}_{seg}$ equally promotes the realistic texture generation of object classes (\eg car, washer classes) and stuff classes (\eg sky, earth classes), the mIoU of object classes (mO) and stuff classes (mS) increase by 2.9 and 3.7,  respectively, which greatly improves the performance of the synthesized images on mIoU and Acc metrics (3.4 and 0.5 higher than Baseline), it still can not promote the generator to synthesize rich structural details. Intuitively, its FID score has been slightly improved.
	(ii) The generator by introducing SPD as an additional condition can synthesize richer details, such as realistic car windows and lights, thereby greatly improving the mIoU, Acc, and FID scores of the generated results \notsure{(see \textit{Baseline VS. Baseline+SPD})}. From Table~\ref{tab:ablation} we can see that the mIoU of the object classes of the \textit{Baseline+SPD} model increased by 3.3 compared to the \textit{Baseline} model, and the FID reduced from 54.2 to 50.1, which indicates the effectiveness of our proposed SPD. 
	(iii) The SAFM block enables the generator to better model the mapping between the SPD features and the shape appearance, which further improves the performance of our model, especially on the mIoU of object classes. 
	(iv) By incorporating the SPD features, SAFM block, and $\mathcal{L}_{seg}$, the performance of \textit{Ours-Full} achieved the best performance, indicating the effectiveness of each component of our method in the synthesizing process.
	
	\begin{table}
		\begin{center}
			\small
			\renewcommand\arraystretch{1}
			\setlength{\tabcolsep}{2.mm}
			{
				\begin{tabular}{c|c|c|c|c|c}
					\hline
					\textbf{Methods} & \textbf{mIoU $\uparrow$} & \textbf{mS $\uparrow$}& \textbf{mO $\uparrow$} & \textbf{Acc $\uparrow$} & \textbf{FID $\downarrow$} \\
					\hline
					\textit{Baseline} & 66.0 & 70.0 &60.7 & 82.5 & 54.2  \\
					+$\mathcal{L}_{seg}$& 69.4& 73.7& 63.6 & 83.0 & 53.2  \\
					\textit{+SPD} & 68.5 & 71.8 & 64.0 & 82.7 & 50.1  \\ 
					\textit{+SPD+SAFM} & 69.4 & 71.5 & \textbf{66.4} & 82.8 & 50.6  \\ 
					\textit{Ours-Full} & \textbf{70.4} & \textbf{74.2} & 65.3 & \textbf{83.1} & \textbf{49.5}  \\
					\hline
				\end{tabular}
			}
		\end{center}
		\vspace{-12pt}
		\caption{Quantitative comparison of five variants on Cityscapes. Here, mS (mO) represents the mIoU of Stuff (object) classes.}
		\label{tab:ablation}
		\vspace{-10pt}
	\end{table}
	
	\subsection{Limitation and Impact}
	\noindent\textbf{Limitation.} Since the part-level semantic layouts for each object class are learning from data, the performance of our approach heavily depends on the quantity of training data. Thus, the rare object classes or the rare shape patterns cannot be well modeled. For instance, non-rigid human bodies sometimes have uncommon posture and shape, from which it is hard to infer the implied part-level layout with insufficient training samples. Nevertheless, our method cloud significantly improve the quality of image synthesis for common object classes and common shape patterns.
	
	\noindent\textbf{Impact.} This paper proposes a method for semantic image synthesis which can synthesize or edit images based on semantic maps. Malicious usage of semantic image synthesis models may have adverse social repercussions, such as the synthesis of images for the purpose of spreading fake news.
	
	\section{Conclusion}
	
	In this paper, the shape of object instances is explicitly encoded into the proposed SPD features. Thus, the object’s part-level layouts could be exploited to improve the generation of images with rich details. Furthermore, the SAFM block is proposed to combine the semantic map and SPD features through conditional convolution operation, which could adaptively modulate the input features. The quantitative and qualitative results demonstrate the superior performance of our method in synthesizing semantically aligned images with rich as well as photo-realistic details.
	
	\noindent\textbf{Acknowledgments.} This work was supported in part by National Key R\&D Program of China under Grant No. 2020AAA0104500, and by the National Natural Science Foundation of China (NSFC) under Grant No.s U19A2073 and 62006064.
	\clearpage
	
	{\small
		\bibliographystyle{ieee_fullname}
		\bibliography{egbib}
	}
	\clearpage
	
	\clearpage
	\section*{Appendix}
		
		\vspace{-15pt}
		\begin{algorithm}
		\caption{Calculation of the SPD}
		\label{alg:1}
		
		\begin{algorithmic}
			\STATE \textbf{Input:} Points set inside the instance object $\textbf{P}\in \mathcal{R}^{m\times 2}$ and points set $\textbf{C}\in \mathcal{R}^{n\times 2}$ on the contour of the instance. Additional input $rbins$ and $tbins$ represent the distance and angle intervals, respectively.
			\STATE \textbf{Output:} The SPD map $\textbf{G}\in \mathcal{R}^{h\times w \times 72}$ of the instance.
			
			\STATE \textit{\textcolor{lightgray}{\# The distance $r$ between points in $\textbf{P}$ and $\textbf{C}$}.} 
			\STATE 1. xdis = \textbf{P[:, 0]}.reshape((-1, 1)) - \textbf{C[:, 0]}.reshape((1,-1))\\
			\quad ydis = \textbf{P[:, 1]}.reshape((-1, 1)) - \textbf{C[:, 1]}.reshape((1,-1))
			\STATE 2. rarray = torch.sqrt(xdis ** 2 + ydis ** 2)\\
			\quad rarray = rarray / (torch.max(rarray) / 2) \\
			
			\STATE \textit{\textcolor{lightgray}{\# The number of points in different distance intervals.}}
			\STATE 3. for  r  in rbins:\\
			\qquad rq += (rarray $<=$ r)
			
			\STATE \textit{\textcolor{lightgray}{\# The angle $t$ between points in $\textbf{P}$ and $\textbf{C}$}.} 
			\STATE 4. tarray = torch.atan2(xgap, -ygap)\\
			\quad tarray = tarray + 2 * math.pi * (tarray $<$ 0)
			\STATE \textit{\textcolor{lightgray}{\# The number of points in different angle intervals.}}
			\STATE 5. tq = (1 +  torch.floor(tarray / (2 * math.pi / tbins)))
			\STATE 6. Count the number of points in each interval with $rq$ and $tq$ and get the final SPD map $\textbf{G}$.
		\end{algorithmic}
	\end{algorithm}	
	\vspace{-15pt}

		\begin{table*}[hb]
		\begin{center}
			\vspace{-5pt}
			\renewcommand\arraystretch{1.}
			\setlength{\tabcolsep}{1.mm}
			{
				\begin{tabular}{c c c c c c c c c c}
					\toprule[1.pt]
					Method & Person & Rider & Car & Truck & Bus & Train & Mcycle &Bcycle & All Objects \\
					\hline
					SPADE & 14.0/62.27 & 16.4/38.67 & 26.6/88.68  & 18.0/64.95 & 28.5/70.16 & 7.9/41.44 & 5.4/28.59 & 9.2/58.86 & 15.7/56.70  \\
					LGGAN & 14.8/64.47 & 18.3/45.99 & 27.4/90.17 & 18.8/73.29 & 34.1/\textcolor{red}{79.05} & 11.8/52.73 & 9.1/39.08 & 11.1/61.38 & 18.2/63.27 \\
					OASIS & 11.3/59.53 & 19.5/\textcolor{blue}{47.96} & 19.2/87.37 & \textcolor{red}{28.0}/62.21 & \textcolor{blue}{36.7}/75.04 & \textcolor{blue}{16.3}/\textcolor{blue}{59.95} & \textcolor{red}{9.9}/\textcolor{red}{48.35} & 9.3/58.40 & 18.8/62.35 \\
					\hdashline
					\textit{Baseline} & 15.2/65.31 & 17.5/43.54 & 28.1/89.97 & 19.2/68.02 & 34.2/72.05 & \textcolor{red}{18.0}/45.14 & 6.7/38.58 & 10.8/62.88 & 18.7/60.69\\
					\textit{Ours w/o $L_{seg}$} & \textcolor{blue}{17.3}/\textcolor{blue}{66.80} & \textcolor{red}{21.0}/46.85 & \textcolor{red}{31.3}/\textcolor{blue}{90.53} & 22.4/\textcolor{red}{78.81} & 34.4/75.97 & 11.5/\textcolor{red}{61.98} & \textcolor{blue}{9.6}/\textcolor{blue}{46.05} & \textcolor{red}{13.6}/\textcolor{blue}{64.11} &\textcolor{blue}{20.1}/\textcolor{red}{66.39} \\
					\textit{Ours-Full} & \textcolor{red}{17.7}/\textcolor{red}{68.04} & \textcolor{blue}{20.5}/\textcolor{red}{49.90} & \textcolor{blue}{31.1}/\textcolor{red}{91.01} & \textcolor{blue}{24.9}/\textcolor{blue}{76.58} & \textcolor{red}{38.9}/\textcolor{blue}{78.53} & 14.1/47.38 & \textcolor{red}{9.9}/45.28 & \textcolor{blue}{13.1}/\textcolor{red}{65.34} & \textcolor{red}{21.3}/\textcolor{blue}{65.26} \\
					\bottomrule[1.pt]
				\end{tabular}
			}
		\end{center}
		\vspace{-19pt}
		\caption{Per-class quantitative comparison of the Detection (AP) / Semantic Segmentation (mIoU) metrics on object classes of Cityscapes. \textcolor{red}{Red} and \textcolor{blue}{blue} indicate
			the best and the second best results, respectively. \textit{Ours w/o $L_{seg}$} is a variant of \textit{Ours-Full} without using $L_{seg}.$ }
		\label{tab:ins}
		\vspace{-10pt}
	\end{table*}
	
	\section*{A. Calculation of the SPD}
	\vspace{0pt}
	The detailed calculation process of the shape-aware position descriptor (SPD) is shown in the pseudo codes Alg. \ref{alg:1}. In this process, we use GPU to speed up the calculation.

	\section*{B. Network Architecture}
	\vspace{-5pt}
	Our generator is mainly composed of the semantic-shape adaptive feature modualtion (SAFM) block. Table \ref{tab:SAFMBlock} shows the details of the SAFM block. The $f_{t-1}$ denotes the features from the previous layer and the $f_{t}$ denotes the features modulated by current SAFM block, respectively. Seg is the input semantic layouts and SPD is the shape-aware position descriptor maps. Conv and Depthwise Conv represent the convolution operation and depthwise convolution, in which convolution kernels are adaptively predicted from semantic layouts. Unfold extracts sliding local blocks from input for conditional convolution operation. Concat denotes the concatenation operation.
	
	\vspace{16pt}
	\begin{table}[H]
		\begin{center}
			\small
			\renewcommand\arraystretch{1.5}
			\setlength{\tabcolsep}{2.7mm}
			{
				\begin{tabular}{c|c|c|c|c|c|c}
					\hline
					\textbf{Input} & $f_{t-1}$ & \multicolumn{3}{c|}{Seg} & \multicolumn{2}{c}{SPD}\\
					\hline
					
					\multirow{9}{*}{\textbf{SAFM}}
					& &  &\multicolumn{2}{c|}{Conv} & Conv &  \\\cline{4-6}
					
					& &  & &Conv & Unfold &  \\\cline{4-6}
					
					&  & Seg & Conv & \multicolumn{2}{c|}{Depthwise Conv} & SPD \\\cline{4-6}
					
					& $f_{t-1}$ &  & \multicolumn{3}{c|}{Depthwise Conv} &  \\\cline{3-7}
					
					&  & \multicolumn{5}{c}{Concat} \\\cline{3-7}
					
					&   & \multicolumn{5}{c}{Conv} \\\cline{3-7}				
					
					&  & \multicolumn{3}{c|}{Conv} & \multicolumn{2}{c}{Conv} \\\cline{2-7}	
					
					&  \multicolumn{4}{c|}{Element-wise Multiply} & \multicolumn{2}{c}{} \\\cline{2-7}	
					
					&   \multicolumn{6}{c}{Element-wise Sum} \\\cline{2-7}	
					
					\hline
					\textbf{output}& \multicolumn{6}{c}{$f_t$} \\
					\hline
				\end{tabular}
			}
		\end{center}
		\vspace{-5pt}
		\caption{Details of the SAFM block.}
		\label{tab:SAFMBlock}
		\vspace{-5pt}
	\end{table}

	\section*{C. More Qualitative Results}
\begin{figure*}[!t]
		\centering		\includegraphics[width=1.\linewidth]{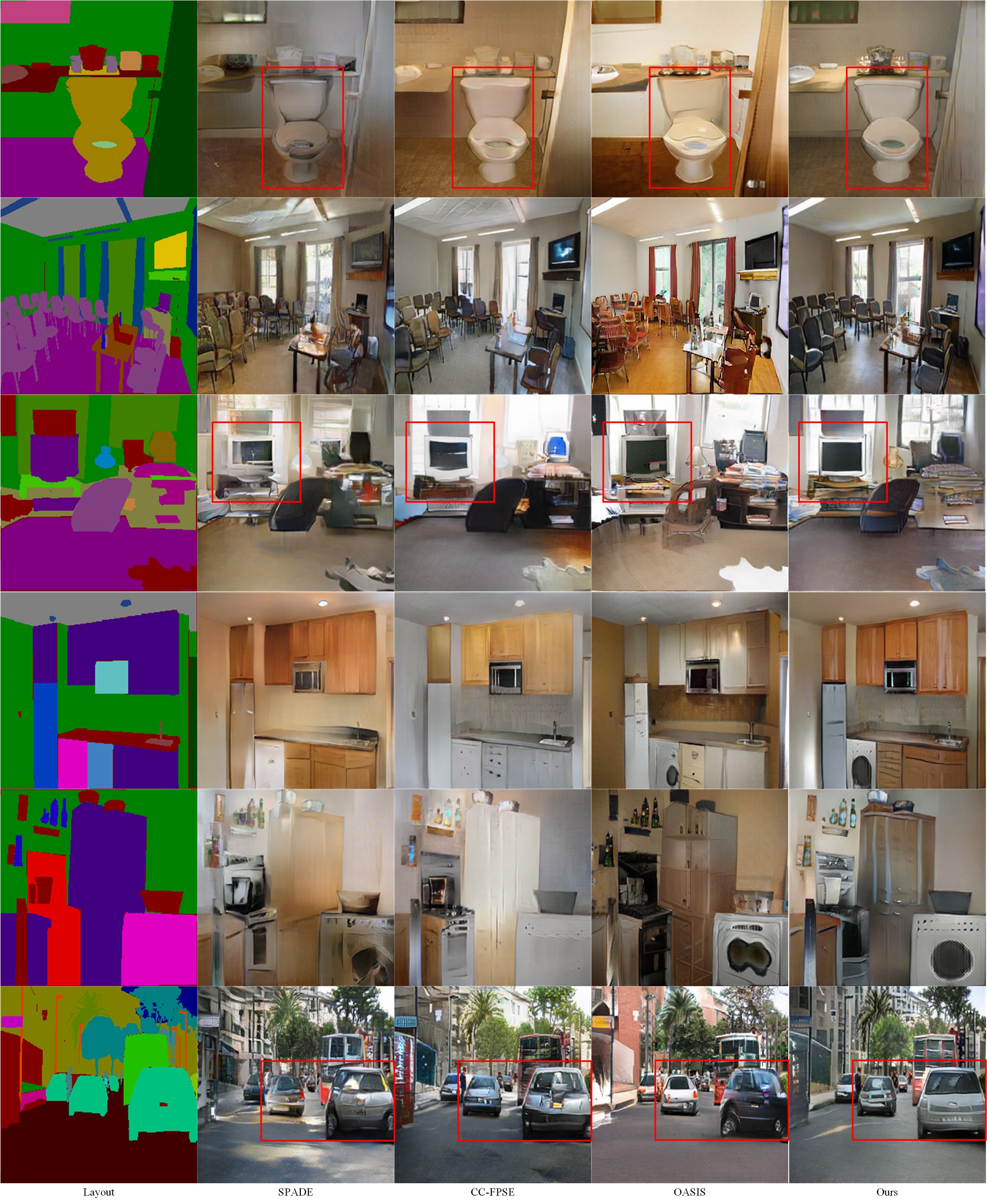}
		\vspace{-18pt}
		\caption{Visual comparisons on the ADE20K and COCO-Stuff datasets.}
		\label{fig:comparea}
		\vspace{-8pt}
	\end{figure*}

\begin{figure*}[!t]
		\centering
		\includegraphics[width=1.\linewidth]{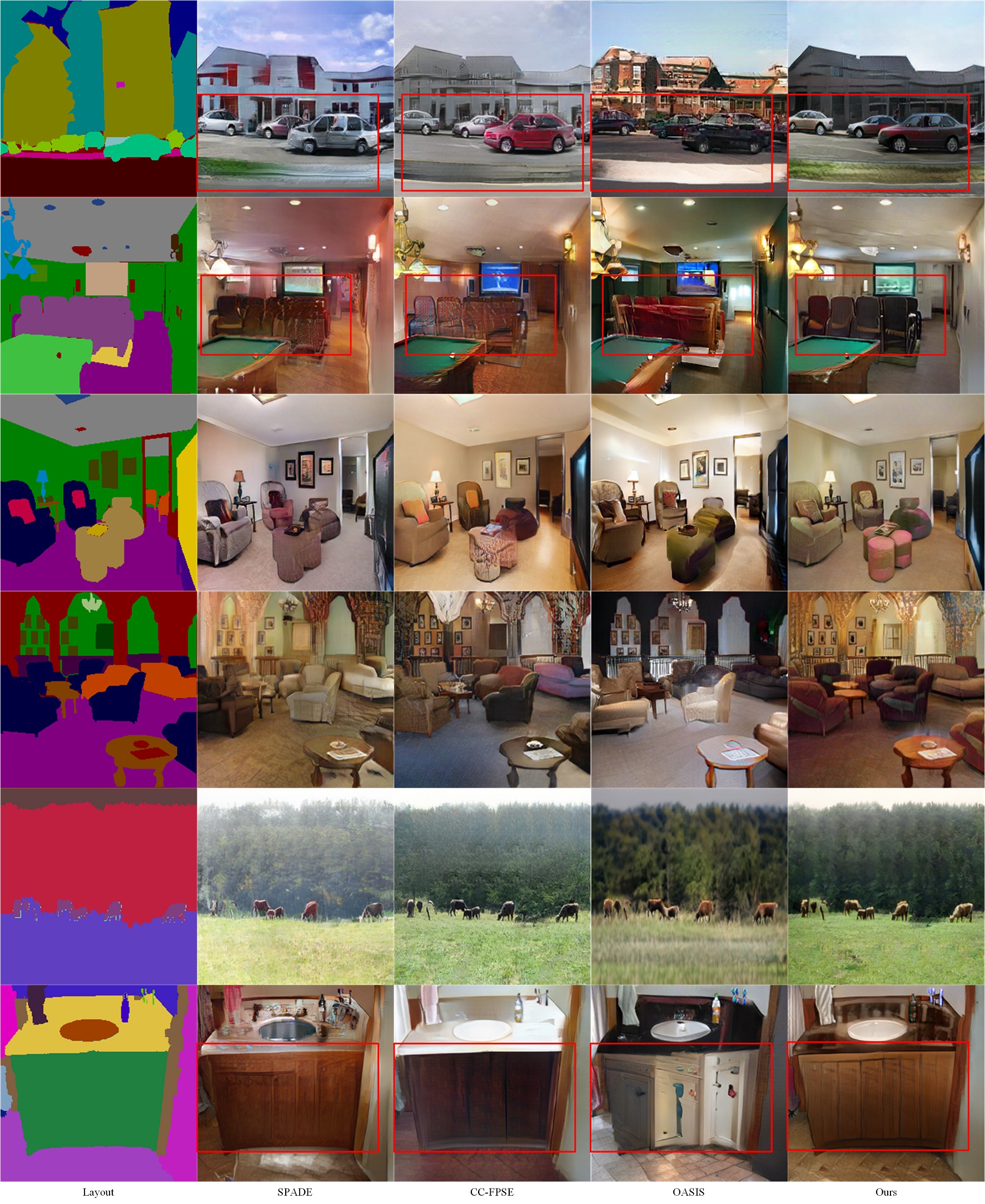}
		\vspace{-18pt}
		\caption{Visual comparisons on the ADE20K and COCO-Stuff datasets.}
		\label{fig:comparec}
		\vspace{-8pt}
	\end{figure*}

	We show more visual comparisons with the competing methods (\ie SPADE~\cite{park2019semantic}, CC-FPSE~\cite{liu2019learning}, and OASIS~\cite{sushko2020you}) on the Cityscapes, ADE20K and COCO-Stuff, as shown in Fig.~\ref{fig:compare}, Fig.~\ref{fig:comparea} and Fig.~\ref{fig:comparec}. Zoom in
	for more details. It can be seen that our method can generate more semantically aligned and photo-realistic images with rich details.
	
\begin{figure*}[!t]
		\vspace{-20pt}
		\centering		\includegraphics[width=.9\linewidth]{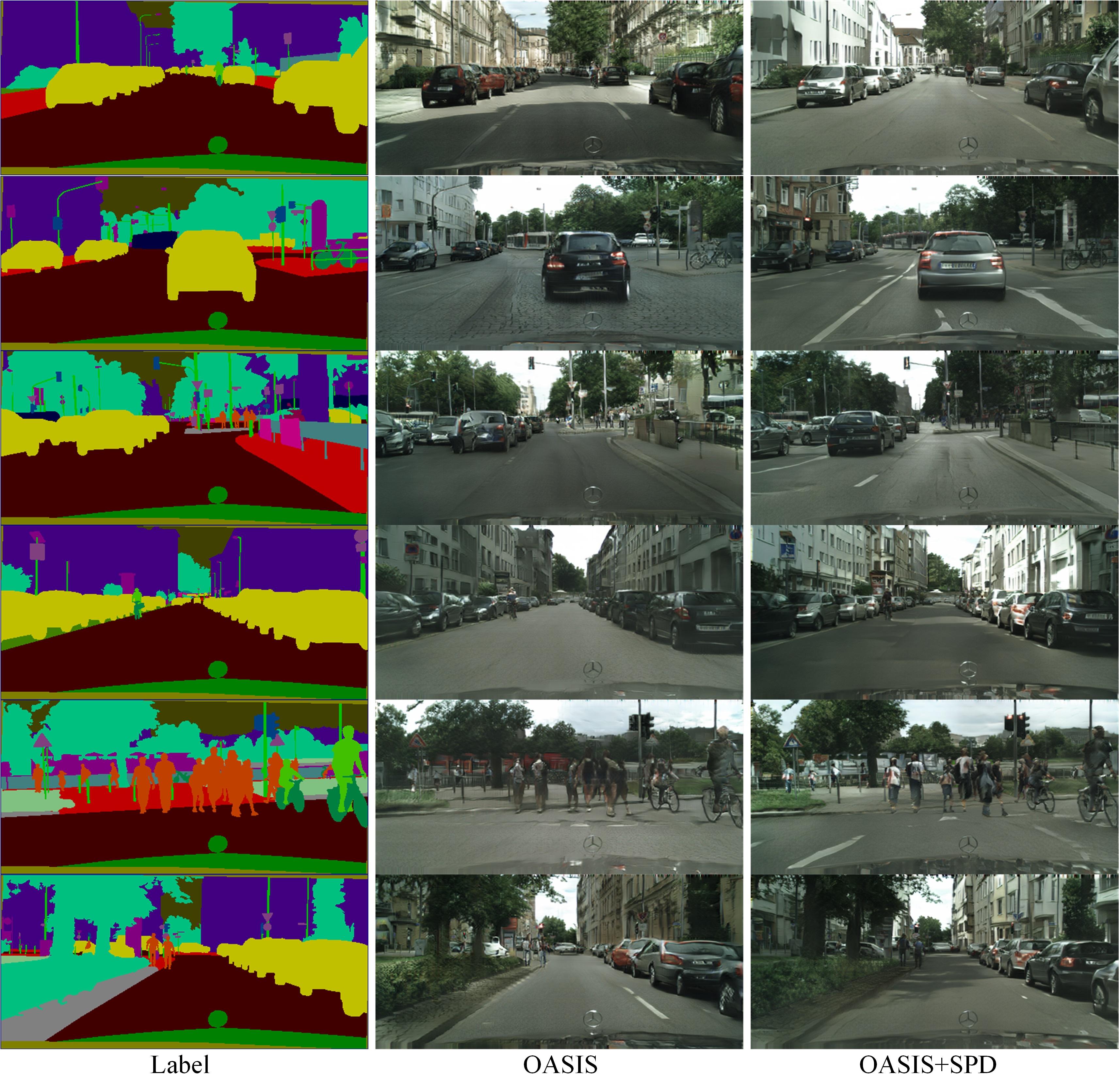}
		\vspace{-13pt}
		\caption{Visual comparisons on the Cityscapes dataset.}
		\label{fig:oasis}
		\vspace{-4pt}
	\end{figure*}

	Additionally, we validate the effectiveness of SPD with OASIS~\cite{sushko2020you} discriminator and training tricks. The generated results are shown in Fig. \ref{fig:oasis}, showing the benefit of SPD for realistic detail synthesis.
	
	\section*{D. More Analysis on Instances Synthesis}
	To quantify the effect of our method on instance synthesis, we evaluate semantic segmentation and detection metrics for instance classes of synthesized results in the Cityscapes dataset.
	The per-class quantitative results of Cityscapes on the synthesized object instance are shown in Table~\ref{tab:ins}.
	One can see that our methods (\textit{Ours w/o $L_{seg}$} and \textit{Ours-Full}) greatly improve the performance upon the \textit{Baseline}, performing favorably against the SoTAs in terms of the semantic segmentation (mIoU) and detection (AP) metrics.
	
	\begin{figure*}[!t]
		\centering
		\includegraphics[width=.95\linewidth]{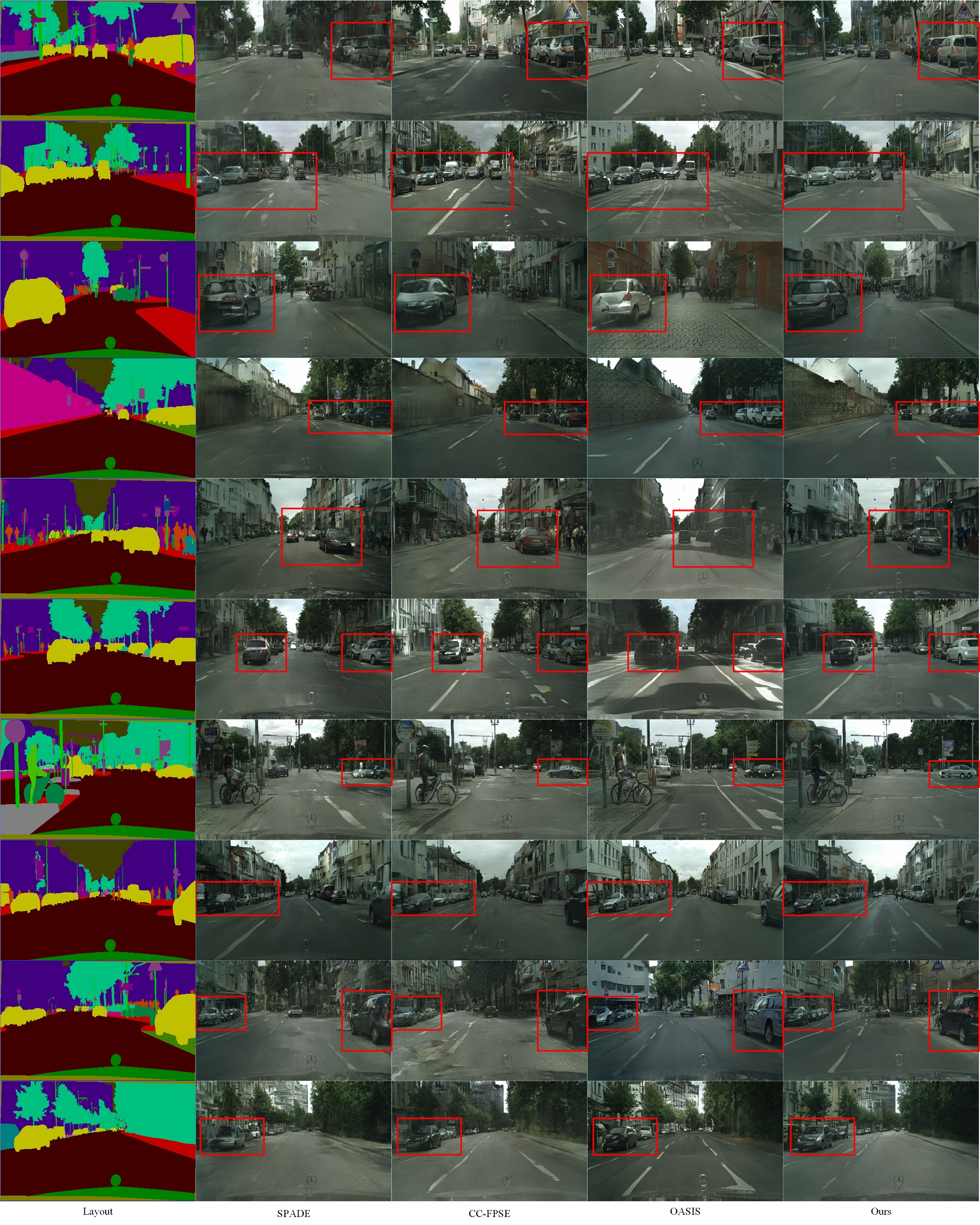}
		\vspace{-8pt}
		\caption{Visual comparisons on the Cityscapes dataset.}
		\label{fig:compare}
		\vspace{-8pt}
	\end{figure*}
	
\end{document}